\documentclass{article}

\usepackage{microtype}
\usepackage{graphicx}
\usepackage{subcaption}
\usepackage{booktabs} 

\usepackage{hyperref}

\usepackage[preprint]{icml2026}

\usepackage{cuted}
\usepackage{multirow}

\usepackage{amsmath}
\usepackage{amssymb}
\usepackage{mathtools}
\usepackage{amsthm}
\usepackage{marvosym}
\usepackage{enumitem}

\usepackage{tcolorbox}
\newtcolorbox{prompt}[1]{
    left=4mm,
    right=4mm,
    top=2mm,
    bottom=2mm,
    boxsep=0mm,
    rounded corners,
    title=#1,
    fontupper=\footnotesize\linespread{0.95}\fontfamily{lmr}\selectfont,
    }
\usepackage[capitalize,noabbrev]{cleveref}

\theoremstyle{plain}

\theoremstyle{definition}

\theoremstyle{remark}

\usepackage[textsize=tiny]{todonotes}

\icmltitlerunning{CoWork-X: Experience-Optimized Co-Evolution for Multi-Agent Collaboration System}

\begin{document}

\twocolumn[
\icmltitle{CoWork-X: Experience-Optimized Co-Evolution for \\ Multi-Agent Collaboration System}


  \icmlsetsymbol{equal}{*}

  \begin{icmlauthorlist}
    \icmlauthor{Zexin Lin}{equal,sse,auto}
    \icmlauthor{Jiachen Yu}{equal,thu}
    \icmlauthor{Haoyang Zhang}{auto}
    \icmlauthor{Yuzhao Li}{auto}
    \icmlauthor{Zhonghang Li}{sse}\\
    \icmlauthor{Yujiu Yang}{thu}
    \icmlauthor{Junjie Wang\textsuperscript{\Letter}}{auto,thu}
    \icmlauthor{Xiaoqiang Ji\textsuperscript{\Letter}}{sse,sai,air}
  \end{icmlauthorlist}

  \icmlaffiliation{sse}{School of Science and Engineering, The Chinese University of Hong Kong, Shenzhen}
  \icmlaffiliation{sai}{School of Artificial Intelligence, The Chinese University of Hong Kong, Shenzhen,}
  \icmlaffiliation{thu}{Tsinghua University}
  \icmlaffiliation{auto}{AutoGame Research}
  \icmlaffiliation{air}{Shenzhen Institute of Artificial Intelligence and Robotics for Society}

  \icmlcorrespondingauthor{Junjie Wang}{wangjunjie@sz.tsinghua.edu.cn}
  \icmlcorrespondingauthor{Xiaoqiang Ji}{jixiaoqiang@cuhk.edu.cn}

  \icmlkeywords{Machine Learning, ICML}

  \vskip 0.3in
]



\printAffiliationsAndNotice{}  

\begin{abstract}
Large language models are enabling language-conditioned agents in interactive environments, but highly cooperative tasks often impose two simultaneous constraints: sub-second real-time coordination and sustained multi-episode adaptation under a strict online token budget. 
Existing approaches either rely on frequent in-episode reasoning that induces latency and timing jitter, or deliver post-episode improvements through unstructured text that is difficult to compile into reliable low-cost execution. 
We propose \textbf{CoWork-X}, an active co-evolution framework that casts peer collaboration as a closed-loop optimization problem across episodes, inspired by fast--slow memory separation. 
CoWork-X instantiates a \textbf{Skill-Agent} that executes via HTN (hierarchical task network)-based skill retrieval from a structured, interpretable, and compositional skill library, and a post-episode \textbf{Co-Optimizer} that performs patch-style skill consolidation with explicit budget constraints and drift regularization. 
Experiments in challenging Overcooked-AI-like realtime collaboration benchmarks demonstrate that CoWork-X achieves stable, cumulative performance gains while steadily reducing online latency and token usage.
\end{abstract}

\section{Introduction}

Large language models (LLMs) accelerate the development of language-conditioned agents in interactive environments and shift communication, planning, and collaboration from handcrafted rules toward a unified learnable paradigm, e.g., in cooperative games. 
In highly interactive cooperative tasks such as Overcooked-AI, this trend brings two simultaneous requirements. 
\textbf{Real-time coordination} requires stable role assignment and reliable handoffs within a sub-second action loop. 
\textbf{Iterative adaptation} requires continuous strategy updates across repeated trials to handle changes in team partners and environment layouts.

\begin{figure}[t]
\centering
\includegraphics[width=0.48\textwidth]{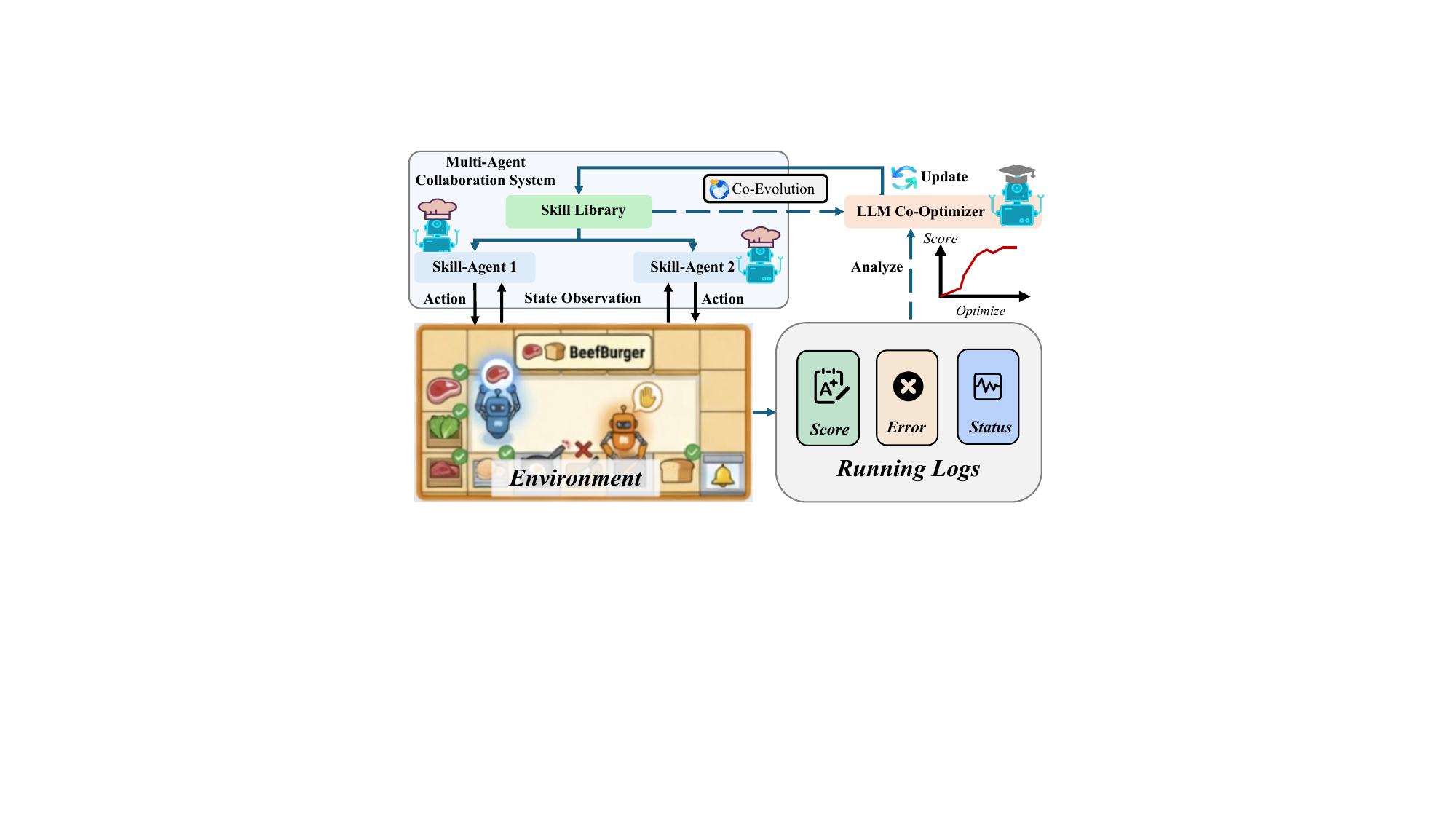} 
\caption{\textbf{CoWork-X overview.} Skill-Agents execute via a shared skill library, and an LLM Co-Optimizer updates it from episode logs for closed-loop co-evolution.}
\vspace{-6mm}
\label{fig:intro}
\end{figure}

To meet these requirements, prior work broadly progresses along three directions. 
(i) \textbf{In-episode reasoning-driven control}: ReAct~\citep{yao2023react} interleaves reasoning and action to improve interactive decision making. 
However, in high-frequency interaction, it triggers frequent reasoning calls. 
These calls introduce online latency and timing jitter, which undermine real-time stability. 
(ii) \textbf{Post-episode language reflection and memory}: Reflexion~\citep{shinn2023reflexion} improves subsequent behavior via after-the-fact reflection on within-episode planning errors. 
Yet its gains are mainly delivered through textual feedback or prompt updates. 
Such textual feedback lacks directly callable, structured skill representations and consistency constraints. 
As a result, experience is not reliably converted into low-cost units for in-episode execution. 
(iii) \textbf{Fast--slow role division, collaboration adaptation, and stronger evaluation pressure}: HLA~\citep{liu2023llmpowered} and DPT-Agent~\citep{zhang-2025-dpt} adopt hierarchical or dual-process designs to ease the conflict between real-time execution and reasoning cost. 
ProAgent~\citep{zhang2024proagent} emphasizes adaptation to unfamiliar teammates. 
Meanwhile, benchmarks such as Collab-Overcooked~\citep{carroll2019utility} and OGC~\citep{sarkar2024ogc} amplify process-level collaboration demands and generalization pressure. 
This setting exposes a hard tension between continuous adaptation and strict constraints on online budget and stability.
In addition, many language-agent paradigms remain \textit{assistant-oriented}, centering on a single LLM serving a human or fixed teammate; fewer studies jointly model all players/co-actors to design a cooperative system that all actors can optimize and co-evolve.

To address these gaps, we propose \textbf{CoWork-X}, an active co-evolution framework for peer multi-agent collaboration. 
CoWork-X shifts the objective from maximizing single-episode performance to achieving \textit{sustained collaboration improvement} across episodes. 
Motivated by the medial temporal lobe declarative memory system~\citep{squire1991medial}, it keeps in-episode behavior fast via rule-like skill memory, and conducts explicit post-episode review to revise and consolidate skills.
Accordingly, CoWork-X follows an \textit{Execute--Optimize} closed loop with an implementation-driven structure. 
(i) \textbf{Skill-Agent (HTN-executable skill memory)}: CoWork-X adopts hierarchical task networks (HTN)~\citep{nau2003shop2} for skill representation, and compiles interpretable, editable, and compositional behaviors into a structured skill library $\mathcal{S}$. 
Within an episode, decision making primarily involves skill retrieval and invocation, reducing reliance on heavyweight online reasoning. 
(ii) \textbf{Co-Optimizer (skill consolidation)}: after each episode, it reads the last trajectory and updates $\mathcal{S}$ based on outcomes and error signals, revising and consolidating skills for future execution. 
(iii) \textbf{Closed-loop iterative optimization}: across episodes, CoWork-X repeats an execute$\rightarrow$diagnose$\rightarrow$update$\rightarrow$re-execute cycle across episodes. 
Expensive reasoning and structural edits are pushed into the update stage, keeping in-episode control lightweight while enabling stable gains over time.

We evaluate CoWork-X on Overcooked-AI under settings that stress process-level collaboration and generalization across teammates and layouts. 
We report cooperative return and online cost (latency, tokens). 
Across $30$ episodes, CoWork-X improves steadily ($52.0$ at $10$ episodes; $96.3$ at $30$), while ReAct/Reflexion/DPT-WToM achieve $12.5/-58.0/3.5$ at $10$ episodes, respectively.
Crucially, CoWork-X executes with $0$ online tokens and $2.6\,s$ per episode, approximately $27\times$ faster than DPT-WToM ($71.0\,s$). 
These results indicate stable gains from closed-loop iteration with sharply reduced in-episode reasoning burden.
Our contributions are:
\begin{itemize}[nosep,leftmargin=*]
\item We propose \textbf{CoWork-X}, an active co-evolution framework for peer multi-agent collaboration that formulates real-time cooperation as a multi-episode closed loop under a strict online budget.
\item We introduce an HTN-based \textbf{Skill-Agent} with a structured \textbf{skill library} for lightweight in-episode execution, and a \textbf{Co-Optimizer} that performs budgeted, regularized patch-style updates for controllable skill evolution.
\item We evaluate CoWork-X under strong collaboration and generalization pressure, showing sustained gains from closed-loop iteration with reduced online cost.
\end{itemize}

\section{Related Work}

\subsection{Real-Time Interactive Collaboration}

Real-time multi-agent coordination under strict latency budgets has inspired diverse architectures. 
ReAct~\citep{yao2023react} shows that explicit reasoning traces can improve performance, but frequent online LLM calls introduce latency and behavioral jitter that break sub-second coordination. 
To reduce in-episode cost, hierarchical designs split fast execution from slow deliberation: HLA~\citep{liu2023llmpowered} uses a fast-mind for reactive actions and a slow-mind for high-level reasoning.
DPT-Agent~\citep{zhang-2025-dpt} further grounds fast control in FSMs with asynchronous reflection and theory-of-mind reasoning. 
ProAgent~\citep{zhang2024proagent} infers teammate intent to coordinate with novel partners, but still relies on online LLM reasoning for both action generation and teammate modeling, incurring latency. 
Despite these advances, prior methods rarely compile experience into structured, verifiable, and compositional skills that support controlled cross-episode evolution, and most remain \textit{assistant-oriented} rather than jointly modeling all actors as peers. 
Benchmarks such as Overcooked-AI~\citep{carroll2019utility} and extensions like OGC~\citep{sarkar2024ogc} intensify these demands via implicit communication, dynamic roles, and zero-shot transfer, while DPT-Agent~\citep{zhang-2025-dpt} further stresses time efficiency.
In contrast, we introduce CoWork-X, a peer co-evolution framework that casts coordination as a multi-episode closed loop. 
It enables stable real-time coordination while consolidating experience into skills that improve controllably across episodes.

\begin{figure*}[!tp]
\centering
\includegraphics[width=\textwidth]{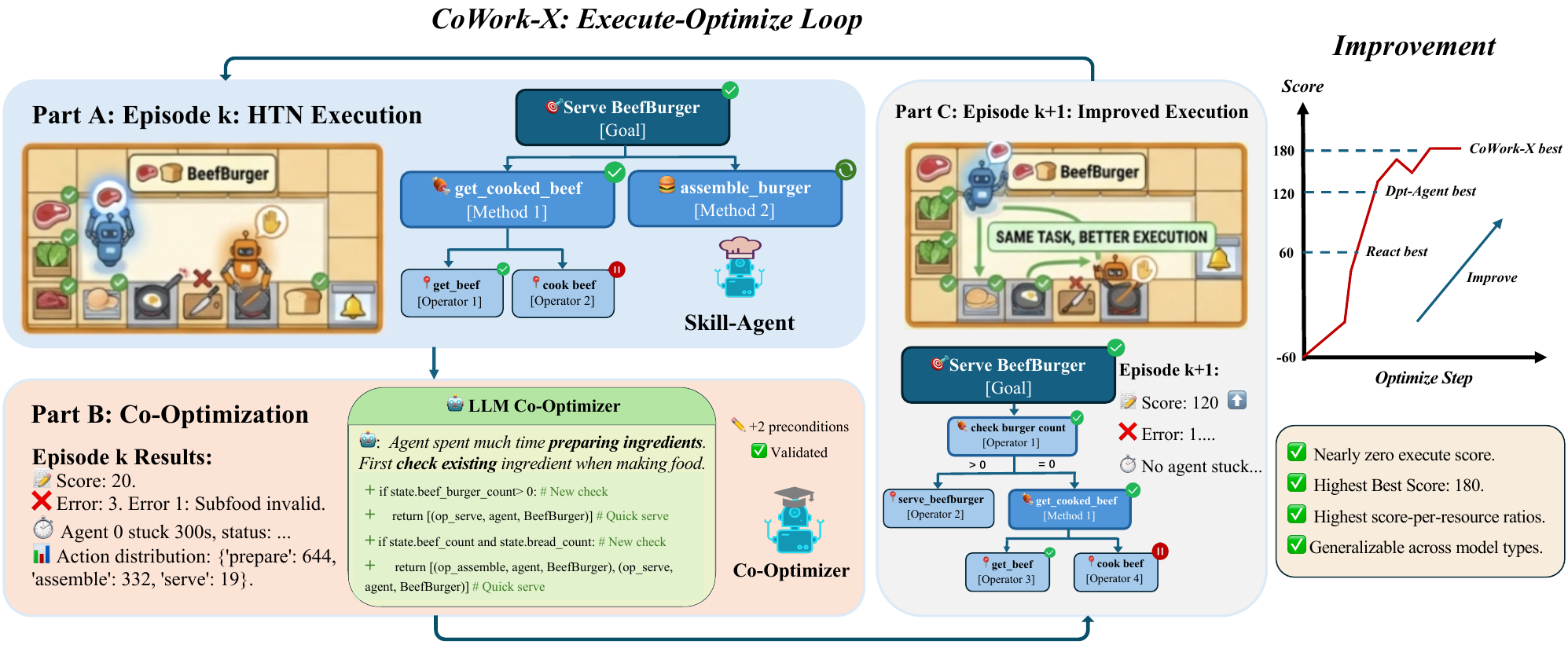} 
\caption{\textbf{CoWork-X Execute--Optimize loop.} A Skill-Agent executes an HTN policy from $\mathcal{S}_k$, then an LLM Co-Optimizer diagnoses episode logs and patches $\mathcal{S}_k\!\to\!\mathcal{S}_{k+1}$ (e.g., adding preconditions), improving performance in subsequent episodes.}
\label{fig:main}
\end{figure*}

\subsection{Slow Adaptation in Multi-Agent Systems}

Sustained multi-agent collaboration requires leveraging past experience to refine coordination over time. 
Multi-agent RL (e.g., MAPPO~\citep{yu2021mappo}) can learn effective joint strategies via centralized training with decentralized execution, but typically yields fixed policies that do not support continual self-improvement or adaptation to non-stationary partners. 
Population-based variants such as COLE~\citep{zhao2023cole} broaden partner coverage, yet they still lack post-deployment skill evolution driven by explicit reasoning. 
Reflection-based agents instead use language feedback for iteration: Reflexion~\citep{shinn2023reflexion} stores textual reflections, and Voyager~\citep{wang2023voyager} accumulates LLM-generated code skills. 
However, these updates are largely prompt or memory driven and often miss structural constraints, verification, and drift regularization needed for controllable accumulation. 
Classical symbolic planning provides verifiability: HTN~\citep{erol1996umcp,nau2003shop2} decomposes goals into primitive actions, and LLM+P~\citep{liu2023llmp} couples LLM understanding with planners for correctness, but the symbolic module is usually treated as a fixed component rather than an evolving repository that closes the loop between execution, diagnosis, and refinement.
Meanwhile, most LLM-agent paradigms remain \textit{assistant-oriented}, where a single agent adapts to a human or fixed partner. 
In contrast, CoWork-X treats all actors as peers sharing a skill library $\mathcal{S}_k$ (iteration-$k$), enabling symmetric co-evolution via budgeted post-episode updates under strict online constraints.

\section{The CoWork-X Framework}
\label{s:cowork-x-framework}

\subsection{Problem Formulation}
\label{ss:problem-setting}

We model peer multi-agent collaboration as a fully observable Decentralized MDP (Dec-MDP) with a meta-learning objective over repeated episodes. 
Formally, the system is a tuple $\langle \mathcal{I}, \mathcal{X}, \{\mathcal{A}^i\}_{i \in \mathcal{I}}, \mathcal{T}, \mathcal{R}, \gamma \rangle$, where $\mathcal{I}=\{1,\ldots,N\}$ indexes agents, $\mathcal{X}$ is the global state, and agent $i$ selects $a^i\in\mathcal{A}^i$. 
The joint action $\mathbf{a}\in\mathcal{A}=\prod_{i=1}^N\mathcal{A}^i$ induces transitions $\mathcal{T}:\mathcal{X}\times\mathcal{A}\to\Delta(\mathcal{X})$, and the team reward is given by $\mathcal{R}:\mathcal{X}\times\mathcal{A}\to\mathbb{R}$ with discount $\gamma\in[0,1)$.

We ground this framework in real-time Overcooked burger preparation scenarios (detailed in \cref{ss:setup}) where $N=2$ agents collaborate to fulfill time-sensitive food orders. The state space $\mathcal{X}$ encodes kitchen layout including ingredient stations, cooking equipment, agent positions, and a queue of pending orders with decreasing time-to-deadline for each active order. Each agent observes the complete state $\mathcal{X}$, though agents must infer teammate intentions from observed actions rather than direct communication. Each agent's action space $\mathcal{A}^i$ includes several atomic operations like get beef or cook beef which are defined in low-level game engine architecture. The reward function $\mathcal{R}$ is controlled by the number of correctly completed orders.

\subsection{Overview: Execute--Optimize Closed Loop}
\label{ss:overview}

As shown in~\cref{fig:main}, our CoWork-X framework reconciles real-time coordination with cross-episode adaptation via an \textit{Execute--Optimize} loop coupled by a skill library $\mathcal{S}_k$. 
$\mathcal{S}_k$ is implemented as a Python file that specifies agent behavior, including state-query utilities, preconditions/procedures for atomic actions, and HTN-encoded task decompositions for diverse orders. 
In \textit{execution mode}, the Skill-Agent retrieves and invokes skills from $\mathcal{S}_k$ for fast rule-based control. 
In \textit{optimization mode}, the Co-Optimizer analyzes previous trajectory $\tau_k$ offline to diagnose recurrent failures and bottlenecks, then updates $\mathcal{S}_k$ by synthesizing patches (e.g., adding missing preconditions or completing recipe logic).

The loop follows $\text{Execute}(\mathcal{S}_k)\!\to\!\text{Diagnose}(\mathcal{D}_k)\!\to\!\text{Update}(\mathcal{S}_k\!\to\!\mathcal{S}_{k+1})\!\to\!\text{Execute}(\mathcal{S}_{k+1})$, repeated over iterations. 
By shifting expensive reasoning to post-hoc offline optimization, CoWork-X keeps online execution lightweight while continuously refining $\mathcal{S}_k$.

\subsection{Skill Agent: HTN-Executable Skill Memory}
\label{ss:skill-agent}

The agent achieves in-episode execution through an HTN planning framework. 
The skill library $\mathcal{S}_k$ consists below three hierarchical functions: 

\begin{itemize}[nosep,leftmargin=*]
\item \textbf{State representation.} Extracts ingredient quantities from the environment and forms a compact state abstraction that omits spatial details (e.g., object/agent positions). This design prioritizes task-relevant features for fast high-level planning, while delegating navigation and manipulation to the midagent controller.
\item \textbf{Operators.} Define atomic food-preparation primitives. Preparation operators update ingredient-count attributes to simulate processing; assembly operators first check ingredient availability, then update raw-material and product counters; serving operators validate finished-dish inventory, decrement counts to fulfill orders and accrue rewards, or return error signals when insufficient.
\item \textbf{Methods.} Map high-level orders to ordered sequences of operators. During execution, a mid-level planner grounds abstract operators into low-level game-engine primitives (e.g., pathfinding and interactions).
\end{itemize}

\subsection{Co-Optimizer: Skill Consolidation and Iterative Updates}
\label{ss:co-optimizer}

The skill library is the primary determinant of the Skill-Agent's game behavior, governing decomposition from high-level objectives to low-level tasks, and is iteratively refined by the Co-Optimizer.

At the onset of the experiment, we initialize $\mathcal{S}_0$ as a minimal but executable HTN scaffold (Pyhop): a small set of core operator/function templates (e.g., \texttt{is\_available}, \texttt{op\_prepare\_food}, \texttt{op\_assemble}, \texttt{op\_serve}) with conservative placeholder logic, plus a few seed methods (e.g., a basic burger routine) and required registrations. 
This ``infant-state'' initialization provides an inductive bias without encoding expert heuristics, while still forcing the Co-Optimizer to infer missing preconditions and state updates, complete recipe logic, and expand task decompositions from execution logs via iterative patches.

After each episode, CoWork-X provides the Co-Optimizer with detailed execution logs beyond the scalar score, including: 
(i) \textbf{Runtime failures} caused by unmet preconditions (e.g., preparing without required ingredients) or invalid variable access. For each failure, we record the exact error location and include the complete environment state and action plans from the two preceding and two succeeding timesteps, which are injected into the prompt to guide repairs to the skill library. 
(ii) \textbf{Stagnation phenomena} flagged when an agent remains inactive for 100 consecutive timesteps. We then provide the environment states and action plans from the first five timesteps of the stagnation interval to help the Co-Optimizer resolve deadlocks. 
(iii) \textbf{Action distribution analysis} reporting the breakdown of action types (e.g., preparation, cooking, serving), which helps identify categories with unusually long durations or low efficiency that warrant optimization. Detailed prompt used in Co-Optimizer can be found in~\cref{sec:co-op-prompt}.

We additionally provide the Co-Optimizer with the current skill library file to enforce basic syntactic constraints and discourage radical edits that violate game logic. 
The best-performing historical library is also included to enable reversion when an update is detrimental. 
The Co-Optimizer then decides whether further optimization is needed; if so, it leverages these diagnostics to refine rules and outputs an updated Python file as the skill library for the next episode.

\section{Experimental Settings}
\label{s:experimental-settings}

\subsection{Environment and Task Configuration}
\label{ss:setup}

We evaluate CoWork-X on real-time overcooked-AI-like burger preparation, building on the DPT-Agent environment~\citep{zhang-2025-dpt}. 
Unlike the original human--AI setting (one LLM assistant for a human player), we extend it to \textbf{symmetric agent--agent collaboration}: both agents are controlled by identical Skill-Agents parameterized by the same skill library $\mathcal{S}_k$, instantiating core challenges of Dec-MDP peer coordination.

We adopt Overcooked since it imposes three constraints that directly stress real-time collaboration: 
(i) \emph{implicit communication}—agents must infer intent from observed actions rather than direct messages; 
(ii) \emph{tight temporal coupling}—sub-second action loops require low-latency coordination; 
(iii) \emph{emergent role assignment}—effective play demands dynamic task allocation and handoffs. 
Together, these properties test whether a method can improve peer coordination consistently across repeated trials.

\textbf{Task specification.}
Agents prepare and serve three dishes: \texttt{BeefBurger} (cooked beef + bread), \texttt{LettuceBurger} (chopped lettuce + bread), and \texttt{BeefLettuceBurger} (cooked beef + chopped lettuce + bread). 
Each episode lasts $T{=}500$ timesteps with up to four concurrent orders whose time-to-deadline decreases over time for each active order, enforcing prioritization under time pressure. 
Rewards are $+20$ per delivered order, $+5$ per intermediate subtask completion (e.g., cooking beef, chopping lettuce), and $-10$ for order failure due to timeout or incorrect delivery. 
Both agents share $\mathcal{S}_k$, ensuring performance reflects coordination rather than fixed roles.

\textbf{Initial state.}
CoWork-X starts from a deliberately impaired but executable skill library: HTN rules are syntactically valid yet semantically incorrect. 
Operators return unchanged states without precondition checks, and methods decompose tasks without verifying availability. 
This yields near-zero initial scores and forces the Co-Optimizer to refine coordination through iterative updates. Details of the skill library's initial state can be found in~\cref{sec:skill-lib-detail}~(\cref{fig:origin-skill}).

\textbf{Evaluation protocol.}
For each method, we run $30$ episodes under a fixed environment configuration with stochastic order generation. 
CoWork-X performs $30$ Execute--Optimize iterations: episode $k$ runs with $\mathcal{S}_k$ to produce $\tau_k$ and diagnostics, followed by one Co-Optimizer update to obtain $\mathcal{S}_{k+1}$. 
The baselines run $30$ episodes independently, without cross-episode skill consolidation. 
We log per-episode return $R_k$, online latency (seconds), and total token usage (prompt + completion across all LLM calls), and report the mean over $30$ episodes without outlier removal.

\subsection{Baseline Methods}
\label{ss:baselines}

We compare CoWork-X against three state-of-the-art LLM-based multi-agent cooperative-control paradigms.

\textbf{ReAct}~\citep{yao2023react} interleaves reasoning and action by generating natural-language traces before issuing executable decisions. 
We invoke the LLM every $25$ timesteps (about $20$ calls per $T{=}500$ episode) with a rolling event buffer ($H{=}1$), and parse JSON task assignments for execution. 
Its main drawback is high online latency from frequent calls and no cross-episode consolidation beyond the prompt window.

\textbf{Reflexion}~\citep{shinn2023reflexion} augments ReAct with a reflective layer. 
A reactive controller runs every $25$ timesteps ($H{=}1$), while a reflection step runs every $75$ timesteps ($H{=}15$) to summarize failures and update textual guidelines appended to subsequent prompts. 
This adds strategic context but still relies on in-episode LLM calls and does not crystallize reusable skills across episodes.

\textbf{DPT-Agent}~\citep{zhang-2025-dpt} adopts a dual-process design with asynchronous fast/slow loops and an optional theory-of-mind (ToM) module. 
The urgent loop triggers every 25 timesteps ($H{=}5$) for immediate assignments, and the reflection loop every 75 timesteps ($H{=}15$) for strategic guidance; an FSM fallback covers common subtasks. 
We evaluate \textbf{DPT-WToM} (with ToM) and \textbf{DPT-WoToM} (without ToM). 
Its limitation is reliance on pre-defined structures and prompt-based updates, with no persistent skill evolution beyond the designed scope.

All baselines utilze the same environment, rewards, and LLM backend (Gemini-3-Pro-Preview-Thinking~\citep{google2025@gemini-3-pro} via an OpenAI-compatible API), with matched sampling (temperature $0.7$, top-p $0.95$, max tokens $4096$). 
Differences stem only from control logic and prompting schedules.
Detailed Settings are in~\cref{as:model-specifications}.

\subsection{Evaluation Metrics}
\label{ss:metrics}

We evaluate methods along $3$ complementary dimensions.

\textbf{Task performance.}
Episode return $R_k$ is the total reward over the 500-timestep episode $k$, combining order deliveries ($+20$), intermediate progress ($+5$), and failures ($-10$). 
We report learning curves $\{R_1,\ldots,R_I\}$ over $I{=}10$ iterations, averaged over ten seeds with standard-error bands.

\textbf{Online efficiency.}
Per-episode latency measures wall-clock runtime including decision overhead (HTN planning for CoWork-X; LLM calls for baselines). 
Token usage sums prompt+completion tokens across all in-episode LLM calls; CoWork-X uses \textbf{0} execution tokens.

\textbf{Amortized cost.}
Offline optimization tokens count post-episode Co-Optimizer usage. 
We also report the break-even episode where CoWork-X's cumulative cost (online + offline) matches baseline cumulative cost (online only).

\subsection{Implementation Details}
\label{ss:implementation}

We adopt Gemini-3-Pro-Preview-Thinking as the backbone of the Co-Optimizer. 
Each trial runs $I{=}10$ Execute--Optimize iterations. 
For each optimization call, we allow up to 3 retries with validation feedback and maintain a history window of the most recent 5 iterations. 
Stagnation is flagged by 100 consecutive inactive timesteps.

\begin{figure}[!tp]
    \centering
    \includegraphics[width=0.48\textwidth]{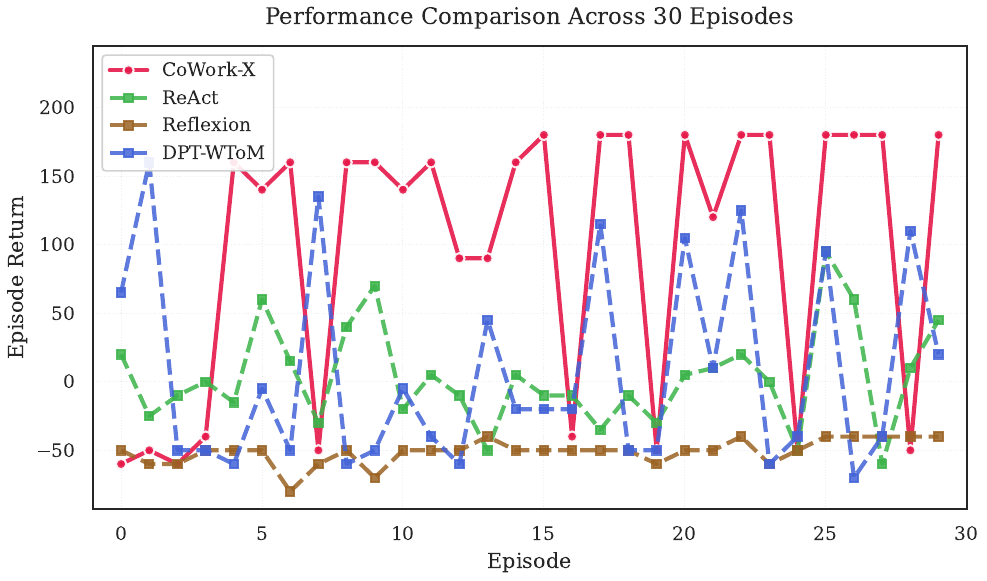}
    \caption{\textbf{Performance across 30 episodes.} CoWork-X shows consistent improvement across iterations. Baselines show unstable performance from frequent online LLM calls.}
    \label{fig:performance_comparison}
\end{figure}

\section{Results and Analysis}

\subsection{Main Results}

\textbf{CoWork-X achieves sustained improvement and remains superior to all baselines throughout the 30-episode trajectory.}
\cref{fig:performance_comparison} shows a clear separation: CoWork-X quickly moves from low initial returns to consistently high performance, and it maintains the top curve for essentially the entire run. 
Importantly, the curve reflects both (i) a positive learning trend across iterations and (ii) persistent dominance over competing approaches, indicating that improvements are not sporadic but repeatedly recovered and maintained. 
In contrast, ReAct and DPT-WToM display high-variance swings with frequent regressions, consistent with online LLM-driven control disrupting sub-second coordination (e.g., delayed action selection causing missed handoffs). 
Reflexion remains largely negative, suggesting that post-hoc textual reflection alone does not capture the fine-grained spatiotemporal contingencies needed for coordinated cooking/serving under time pressure.

\textbf{Block-level aggregation quantifies monotonic gains for CoWork-X and exposes baseline stagnation and instability.}
\cref{tab:episode-range-scores} confirms progressive refinement: CoWork-X improves from $52.0$ (episodes 0--9) to $109.0$ (10--19) to $128.0$ (20--29), with an overall mean of $96.3$. 
This monotonic rise indicates that cross-episode updates consistently consolidate useful coordination logic into the HTN library rather than overfitting to a single episode. 
Baselines fail to exhibit a similar pattern. ReAct oscillates around zero ($3.2$ overall) and even degrades in the middle block ($-16.5$), aligning with the hypothesis that frequent reasoning calls incur latency/jitter that intermittently breaks coordination. 
DPT-WToM shows modest late gains ($25.5$ in 20--29) but remains low on average ($6.2$ overall), implying limited adaptation beyond its fixed policy scaffold despite added ToM overhead. 
Reflexion stays strongly negative across all blocks ($-58.0$, $-50.0$, $-45.0$; $-51.0$ overall), indicating that prompt-level reflections do not reliably translate into executable coordination improvements. 
Detailed per-episode scores are provided in~\cref{sec:raw-results}.

\begin{table}[!tp]
\centering
\caption{Mean performance by episode range (transposed). CoWork-X improves across 10-episode blocks, while baselines remain stagnant or unstable.}
\label{tab:episode-range-scores}
\small
\setlength{\tabcolsep}{5pt}
\begin{tabular}{lcccc}
\toprule
Method & 0--9 & 10--19 & 20--29 & Overall (0--29) \\
\midrule
ReAct      & 12.5  & -16.5 & 13.5  & 3.2 \\
Reflexion  & -58.0 & -50.0 & -45.0 & -51.0 \\
DPT-WToM   & 3.5   & -10.5 & 25.5  & 6.2 \\ \midrule
CoWork-X   & \textbf{52.0} & \textbf{109.0} & \textbf{128.0} & \textbf{96.3} \\
\bottomrule
\end{tabular}
\end{table}

\begin{table}[!tp]
\centering
\caption{Computational costs per episode. CoWork-X uses symbolic planning (zero online tokens); optimization costs are amortized across episodes. All measurements report mean values across all episodes.}
\label{tab:efficiency}
\small
\setlength{\tabcolsep}{4pt}
\begin{tabular}{lrrrr}
\toprule
Method & \multicolumn{2}{c}{Online} & \multicolumn{2}{c}{Total (+ optimization)} \\
\cmidrule(lr){2-3} \cmidrule(lr){4-5}
       & Time (s) & Tokens & Time (s) & Tokens \\
\midrule
ReAct      & 182.3 & 79,126 & 182.3 & 79,126 \\
Reflexion  & 67.7  & 35,635 & 67.7  & 35,635 \\
DPT-WToM   & 71.0  & 30,090 & 71.0  & 30,090 \\ \midrule
CoWork-X   & \textbf{2.6} & \textbf{0} & 160.3 & 22,117 \\
\bottomrule
\end{tabular}
\end{table}

\subsection{Efficiency and Cost Profile}
\label{ss:efficiency}

We measure per-episode cost in a 30-episode Overcooked evaluation, reporting (i) \emph{online} runtime and tokens incurred during episode execution, and (ii) \emph{total} cost that additionally includes post-episode optimization. 

\textbf{CoWork-X nearly eliminates online cost, shifting computation to offline optimization.}
\cref{tab:efficiency} shows that CoWork-X executes with $2.6s$ online time and \textbf{0} online tokens, since the Skill-Agent uses HTN-based symbolic planning without in-episode LLM calls. 
In contrast, all baselines incur substantial online overhead from frequent LLM invocations, ranging from 67.7--182.3s and 30k--79k tokens per episode.

\textbf{Even with optimization included, CoWork-X remains token-efficient while delivering real-time deployment feasibility.}
Including post-episode updates, CoWork-X's total cost is $160.3s$ and $22,117$ tokens per episode on average. Notably, its total token usage is still lower than DPT-WToM ($22.1$k vs $30.1$k), despite paying an explicit optimization budget. Meanwhile, CoWork-X's online latency is over an order of magnitude smaller than baselines (e.g., $2.6s$ vs $71.0s$ for DPT-WToM), making it compatible with sub-second action loops where online LLM calls can systematically break coordination.

\begin{figure}[!t]
\centering
\includegraphics[width=\linewidth]{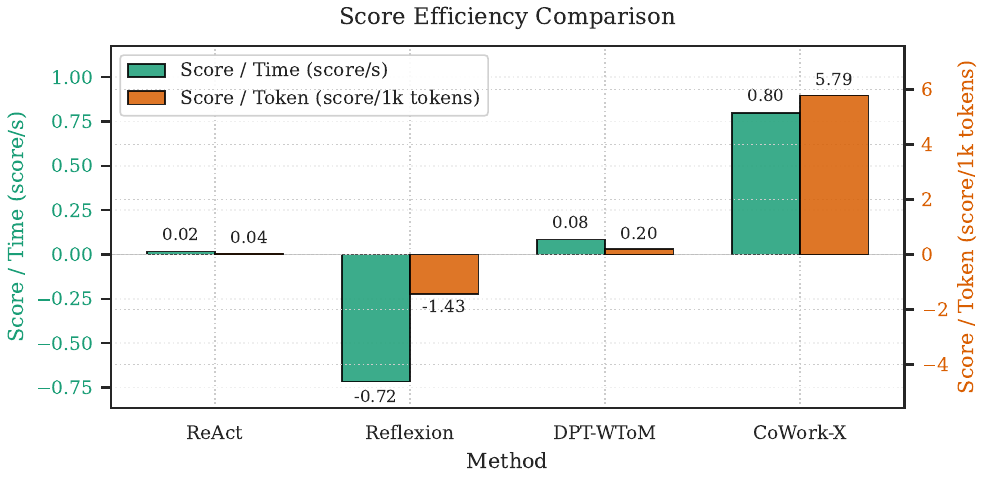}
\caption{\textbf{Score efficiency.} CoWork-X achieves higher score-per-resource ratios: 0.92 score/s and 5.9 score/1k tokens, versus DPT-WToM's 0.09 and 0.20.}
\label{fig:score_efficiency}
\end{figure}

\begin{figure*}[!tp]
    \centering
    \includegraphics[width=\textwidth]{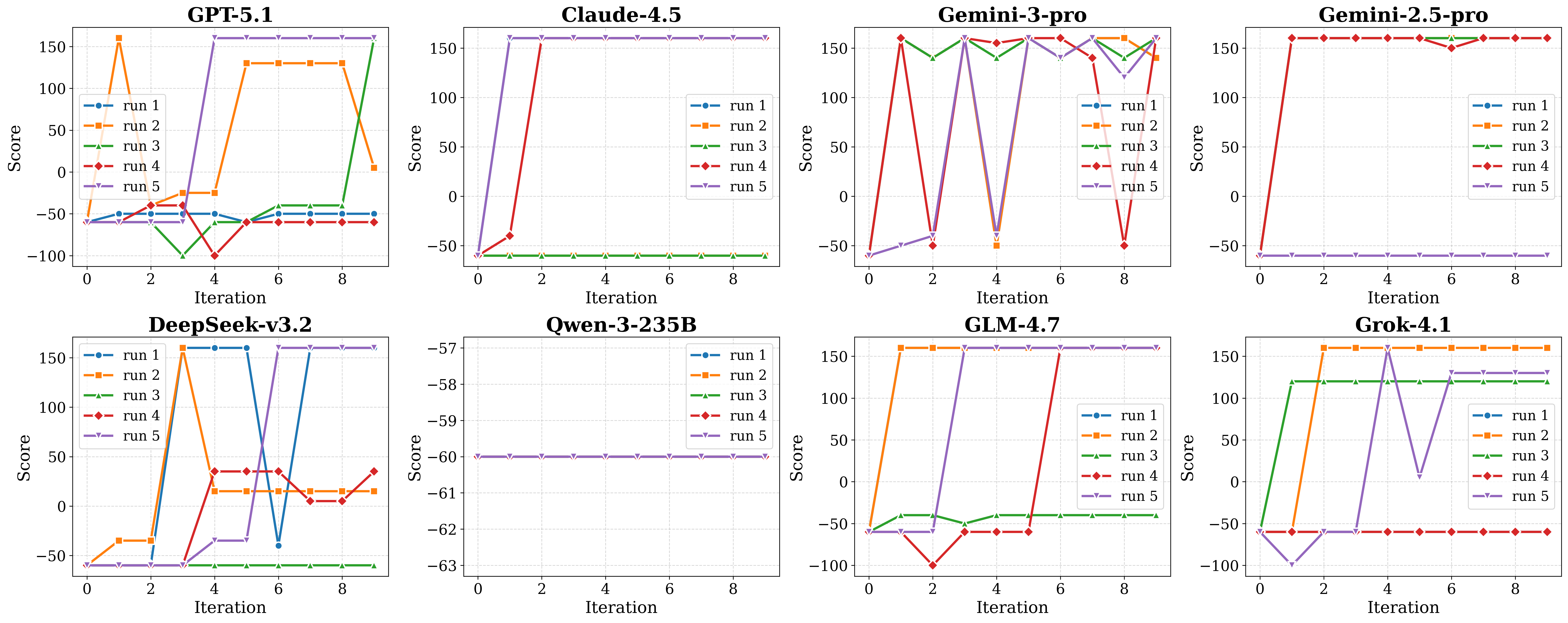} 
    \caption{Ablation study of different families of models. Each model was tested in 5 independent runs, each with 10 iterations.}
    \label{fig:model_ablation}
\end{figure*}

\subsection{Score Efficiency under Online Budgets}
\label{ss:score-efficiency}

We measure \emph{score efficiency} by normalizing average episode return with (i) total wall-clock time and (ii) total token usage, using the same $30$-episode evaluation protocol and mean costs/returns aggregated across episodes. 
\cref{fig:score_efficiency} shows that CoWork-X attains the highest score/time and score/token ratios, achieving $0.92$ score/s and $5.9$ score/1k tokens. 
In comparison, the strongest baseline (DPT-WToM) reaches only $0.09$ score/s and $0.20$ score/1k tokens, while ReAct is near-zero and Reflexion is negative due to consistently low returns.

\textbf{Eliminating online LLM calls is the key driver of real-time efficiency.}
CoWork-X's advantage stems from shifting computation offline: the HTN-based Skill-Agent executes with zero online tokens and minimal latency, while the Co-Optimizer's cost is amortized across episodes. 
This separation yields high returns without incurring the online jitter of frequent LLM calls, making CoWork-X compatible with sub-second coordination loops where latency directly translates into missed handoffs and reduced team reward.

\subsection{Cost Amortization and Deployment Scalability}
\label{ss:amortization}

\begin{table}[!tp]
\centering
\caption{Cost comparison at different episode counts. All values report cumulative mean across the first N episodes. CoWork-X's performance and efficiency advantages increase as optimization costs are amortized.}
\label{tab:amortization}
\small
\setlength{\tabcolsep}{3.5pt}
\begin{tabular}{lrrrrr}
\toprule
Range & Method & Time & Tokens & Mean & Score Per \\
    &        & (s)  &        & Score & 1k Tokens \\
\midrule
\multirow{4}{*}{0-9}
& ReAct & 1,823 & 791,259 & 12.5 & 0.016 \\
& Reflexion & 677 & 356,348 & -58.0 & -0.163 \\
& DPT-WToM & 710 & 300,898 & 3.5 & 0.012 \\
& CoWork-X & 1,603 & 221,166 & \textbf{52.0} & \textbf{0.235} \\
\midrule
\multirow{4}{*}{0-19}
& ReAct & 3,646 & 1,582,518 & -2.0 & -0.001 \\
& Reflexion & 1,354 & 712,695 & -54.0 & -0.076 \\
& DPT-WToM & 1,420 & 601,795 & -3.5 & -0.006 \\
& CoWork-X & 3,206 & 442,332 & \textbf{80.5} & \textbf{0.182} \\
\midrule
\multirow{4}{*}{0-29}
& ReAct & 5,469 & 2,373,777 & 3.2 & 0.001 \\
& Reflexion & 2,031 & 1,069,043 & -51.0 & -0.048 \\
& DPT-WToM & 2,130 & 902,693 & 6.2 & 0.007 \\
& CoWork-X & 4,808 & 663,499 & \textbf{96.3} & \textbf{0.145} \\
\bottomrule
\end{tabular}
\end{table}

In~\cref{tab:amortization}, we report cumulative mean statistics over the first $N\in{10,20,30}$ episodes (ranges 0--9, 0--19, 0--29). For each method, we aggregate wall-clock time, total tokens, mean score, and score-per-1k-tokens. CoWork-X follows an Execute--Optimize schedule with one post-episode update per episode, whereas baselines incur only in-episode costs.

\textbf{CoWork-X maintains a large and early performance lead throughout deployment.}
The result shows that CoWork-X is already strong after $10$ episodes (mean score $52.0$), and the advantage widens with more episodes ($80.5$ at $20$; $96.3$ at $30$). Baselines remain near zero or negative on average over the same horizons (e.g., DPT-WToM: $3.5$, $-3.5$, $6.2$; Reflexion: $-58.0$, $-54.0$, $-51.0$), indicating little systematic learning from additional trials.

\textbf{CoWork-X achieves higher performance at lower cumulative token cost than the strongest baseline.}
By $30$ episodes, CoWork-X reaches $96.3$ using $663$k tokens, while DPT-WToM reaches $6.2$ using 903k tokens. 
This yields a $15.6\times$ score gain with $27\%$ fewer tokens, showing that optimization overhead does not offset overall efficiency.

\textbf{Cost amortization improves CoWork-X's score-per-token profile as episodes accumulate.}
CoWork-X's score-per-1k-tokens decreases from $0.235$ (0--9) to $0.18$2 (0--19) to $0.145$ (0--29), consistent with amortizing a fixed optimization cost over more deployments while sustaining high returns. Baselines show no comparable efficiency trend (e.g., DPT-WToM: $0.012$ $\rightarrow$ $-0.006$ $\rightarrow$ $0.007$), suggesting they primarily accumulate cost without converting experience into reusable coordination capability.

\subsection{Generalization across LLM Backbones}
\label{ss:model-ablation}

We instantiate CoWork-X with diverse LLM backbones spanning 7 model families (GPT~\citep{openai2025@gpt-5}, Gemini~\citep{google2025@gemini-3-pro}, Claude~\citep{anthropic2025@claude-4.5}, DeepSeek~\citep{liu2025deepseek}, Qwen~\citep{yang2025qwen3}, GLM~\citep{@glm}, Grok~\citep{xai2025@grok-4}). 
For each backbone, we run $5$ independent trials, each with $10$ Execute--Optimize iterations, and report score trajectories in~\cref{fig:model_ablation}.

\textbf{CoWork-X generalizes across most backbones, yielding substantial gains from closed-loop optimization.}
Across closed-source families, CoWork-X frequently reaches high scores (often $\ge 140$) in multiple runs, indicating that diverse backbones can synthesize and refine executable HTN skills under the same optimization protocol. 
In contrast, Qwen-3-235B remains near a low-score regime (around $-60$), revealing a backbone-dependent limit in reliable skill repair and refinement. The failure case can be found in~\cref{sec:fail}.

\textbf{Backbones differ in optimization speed and attainable ceilings.}
Claude-4.5 and Gemini-2.5-pro typically reach a high ceiling (around $160$) within roughly $3$ iterations in most runs, whereas GPT-5.1 and DeepSeek-v3.2 improve more slowly and often plateau earlier. 
GLM-4.7 and Grok-4.1 can match top-end ceilings in some runs, but generally require more iterations to reach a stable optimum.

\textbf{Stability varies both within-run and across runs, exposing different failure--recovery profiles.}
Some backbones (e.g., Claude-4.5, Gemini-2.5-pro, GLM-4.7) remain stable once a strong library is found, while others (e.g., Gemini-3-pro, GPT-5.1, DeepSeek-v3.2) exhibit larger oscillations, consistent with occasional harmful patches followed by rollback-driven recovery. 
With ``effective optimization'' defined as stabilizing above $120$ points, the Gemini backbones and GLM-4.7 achieve the highest success rate ($4/5$), while GPT-5.1 and DeepSeek-v3.2 are less consistent ($2/5$).

\begin{figure}[!tp]
    \begin{prompt}{Example of Skill Correction}
{\color{blue} \textbf{Original Rule: }}...if getattr(state, \textbf{plate\_count}) = 0:\\
subtasks.append((op\_prepare\_food, agent, \textbf{Plate}))...\\
{\color{blue} \textbf{Error Log: }}...assert food in [Lettuce, Beef, Bread], food; \textbf{AssertionError: Plate}...\\
{\color{blue} \textbf{Updated Skill: }}\# Plate availability is assumed implicitly...\textbf{delete item Plate}.\\
{\color{blue} \textbf{Score: }}\textbf{-40 → 100}
\end{prompt}
    \caption{ An example of skill correction.
    \label{fig:skill-correction}
    }
\end{figure}

\begin{figure}[!tp]
    \begin{prompt}{Example of Skill Improvement}
{\color{blue} \textbf{Original Rule: }}def is\_available(state, ingredient):\\ if ingredient = Bread:\\
        \textbf{\hspace*{2em} return getattr(state, bread\_count) \textgreater ~0}...\\
{\color{blue} \textbf{Error Log: }}Agent 1: Failed, lack of necessary ingredients to assemble BeefLettuceBurger \textbf{(ingredients may be used by your partner)}...\\
{\color{blue} \textbf{Updated Skill: }}def is\_available(state, ingredient):\\ if ingredient = Bread:\\
\# Need 2 to avoid failure when partner grab one.\\
\textbf{\hspace*{2em} return getattr(state, bread\_count) \textgreater ~1}\\
{\color{blue} \textbf{Score: }}\textbf{160 → 180}
\end{prompt}
    \caption{ An example of skill improvement.
    \label{fig:skill-improvement}
    }
\end{figure}

\subsection{Qualitative Skill Evolution Analysis}

We inspect two representative optimization traces produced by the Co-Optimizer (Gemini-3-pro) when updating the HTN-based skill library from episode logs, and visualize the corresponding rule edits, error signals, and score changes in~\cref{fig:skill-correction,fig:skill-improvement}.

\textbf{Log-grounded patches can repair hard execution failures by removing invalid symbolic assumptions.}
\Cref{fig:skill-correction} shows a typical early-stage failure where the Co-Optimizer introduces a non-existent variable (\texttt{plate\_count}) and an unsupported ingredient (\texttt{Plate}), triggering an \texttt{AssertionError}. 
Using the error location and runtime context, the next update deletes the invalid item and restores consistency with the environment's symbol set, converting a failing trajectory (score $-40$) into a successful run (score $100$). 
This illustrates that CoWork-X is not merely ``prompt tuning'': it performs concrete, verifiable code repairs that directly unblock execution.

\textbf{Beyond crash fixes, the Co-Optimizer refines coordination logic by diagnosing implicit interaction bugs.}
\Cref{fig:skill-improvement} demonstrates a higher-level improvement where execution does not crash but coordination degrades due to resource contention. 
The original availability check for bread (\texttt{bread\_count} $>0$) allows both agents to compete for the last unit, inducing deadlock and downstream order failure. The Co-Optimizer resolves this by tightening the condition to require \texttt{bread\_count} $>1$, effectively encoding a coordination-aware precondition that accounts for teammate consumption. 
The score increase (from $160$ to $180$) highlights that CoWork-X can synthesize non-trivial multi-agent constraints from feedback logs even without explicit fault localization.

\textbf{Rollback enables recovery from harmful edits and stabilizes multi-step evolution.}
In later iterations, the Co-Optimizer may propose overly restrictive rules (e.g., ``burger ownership'' constraints that prevent efficient handoffs), which can reduce score despite being syntactically valid. 
CoWork-X mitigates such regressions by allowing the Co-Optimizer to revert to a previously best-performing library, preventing error accumulation from collapsing the optimization trajectory. 
This failure--recovery pattern complements the quantitative stability trends and is essential for reliable cross-episode co-evolution.

\section{Conclusion}

We presented \textbf{CoWork-X}, an active co-evolution framework for peer multi-agent collaboration that reconciles sub-second coordination with cross-episode adaptation under strict online budgets. 
CoWork-X implements an \textit{Execute--Optimize} loop: Skill-Agents execute lightweight, HTN-structured behaviors from a shared skill library, while an LLM Co-Optimizer performs post-episode diagnosis and patch-style updates from execution logs to consolidate reusable coordination skills. 
Experiments on real-time overcooked-AI-like benchmark demonstrate sustained improvement and strong efficiency. 
Across $30$ episodes, CoWork-X steadily increases return (from $52.0$ at $10$ episodes to $96.3$ at $30$) while baselines remain low or unstable. 
By eliminating in-episode LLM calls, CoWork-X achieves $0$ online tokens and $2.6$s per episode, enabling reliable coordination under tight latency constraints. 
Overall, CoWork-X shows that structured, log-grounded skill evolution can yield scalable and robust peer collaboration beyond prompt-only adaptation.

\section*{Impact Statements}

This work introduces CoWork-X, a closed-loop framework for peer multi-agent collaboration that compiles cross-episode experience into an executable HTN skill library via post-episode LLM-driven updates. 
The intended impact is to improve coordination reliability under strict online latency budgets, which can benefit real-time interactive systems such as assistive robotics, collaborative agents, and human-in-the-loop decision support.

Potential risks arise from automated skill evolution. 
Post-episode code updates may introduce brittle heuristics, reward hacking, or silent regressions that only surface under distribution shift (e.g., new layouts or non-stationary partners). 
LLM-generated patches can also contain implementation bugs or security-relevant flaws, and improved coordination efficiency could be misused to amplify harmful collective behaviors in adversarial settings.

We mitigate these concerns by separating online execution from offline optimization, restricting changes to structured post-episode patches rather than in-episode free-form reasoning, validating generated skill libraries before execution, and keeping a best-performing historical library to support rollback after harmful edits. We also report both performance and cost to make deployment trade-offs explicit. 
Future work should strengthen safeguards with automated testing and invariant checks, sandboxed execution, explicit safety constraints during optimization, and broader evaluation under adversarial and out-of-distribution conditions.

To support reproducibility and responsible reuse, we release our implementation under the MIT License. 
We encourage users to follow the license terms (e.g., attribution and preserving the license notice). 
We also encourage publicly sharing derivatives and results built on CoWork-X to improve transparency, comparability, and collective progress.

\bibliography{ref}
\bibliographystyle{icml2026}

\clearpage
\appendix
\section*{Appendix}

\section{Raw Experimental Results}
\label{sec:raw-results}

Table~\ref{tab:episode-scores} presents the complete episode-by-episode scores from the single experimental run of 30 episodes for each method. CoWork-X demonstrates progressive improvement through iterative skill library refinement, while baselines show stagnant or unstable performance due to lack of cross-episode learning mechanisms.
\begin{table}
\centering
\caption{Episode-by-episode scores for single experimental run. CoWork-X shows consistent improvement across iterations while baselines remain stagnant.}
\label{tab:episode-scores}
\small
\setlength{\tabcolsep}{4pt}
\begin{tabular}{rccccc}
\toprule
Episode & ReAct & Reflexion & DPT-WToM & CoWork-X \\
\midrule
0  & 20  & -50 & 65  & -60 \\
1  & -25 & -60 & 160 & -50 \\
2  & -10 & -60 & -50 & -60 \\
3  & 0   & -50 & -50 & -40 \\
4  & -15 & -50 & -60 & 160 \\
5  & 60  & -50 & -5  & 140 \\
6  & 15  & -80 & -50 & 160 \\
7  & -30 & -60 & 135 & -50 \\
8  & 40  & -50 & -60 & 160 \\
9  & 70  & -70 & -50 & 160 \\
10 & -20 & -50 & -5  & 140 \\
11 & 5   & -50 & -40 & 160 \\
12 & -10 & -50 & -60 & 90  \\
13 & -50 & -40 & 45  & 90  \\
14 & 5   & -50 & -20 & 160 \\
15 & -10 & -50 & -20 & 180 \\
16 & -10 & -50 & -20 & -40 \\
17 & -35 & -50 & 115 & 180 \\
18 & -10 & -50 & -50 & 180 \\
19 & -30 & -60 & -50 & -50 \\
20 & 5   & -50 & 105 & 180 \\
21 & 10  & -50 & 10  & 120 \\
22 & 20  & -50 & 125 & 180 \\
23 & 0   & -40 & -60 & 180 \\
24 & -50 & -60 & -40 & -50 \\
25 & 95  & -50 & 95  & 180 \\
26 & 60  & -40 & -70 & 180 \\
27 & -60 & -40 & -40 & 180 \\
28 & 10  & -40 & 110 & -50 \\
29 & 45  & -40 & 20  & 180 \\
\bottomrule
\end{tabular}
\end{table}

\textbf{Key observations:}
\begin{itemize}
    \item \textbf{CoWork-X progressive improvement}: Scores increase from 52.0 (episodes 0--9) to 109.0 (episodes 10--19) to 128.0 (episodes 20--29), demonstrating effective skill library optimization across iterations. Summary statistics are presented in the main text (\cref{tab:episode-range-scores}).
    \item \textbf{Baseline stagnation}: ReAct, Reflexion, and DPT-WToM show no consistent improvement across episode ranges, as they lack cross-episode learning mechanisms.
    \item \textbf{Final convergence}: CoWork-X achieves 180 points in 7 of the final 10 episodes (episodes 20--29), indicating convergence to effective coordination strategies.
    \item \textbf{Occasional failures}: CoWork-X shows occasional negative scores (-50 to -60) even in later episodes, typically occurring when the Co-Optimizer attempts exploratory modifications that temporarily degrade performance before recovery in subsequent iterations.
\end{itemize}

\section{Prompt of Co-Optimizer}
\label{sec:co-op-prompt}
\cref{fig:optimize-prompt} shows the prompt used by Co-Optimizer. The prompt context consists the current skill library, the origin library, historical performance metrics, and diagnostic error logs. It requires the Co-Optimizer to refine the skill library by addressing the specific runtime errors identified in the previous iteration, with the final output strictly constrained to Python format.

\section{Model Specifications}
\label{as:model-specifications}

\begin{table}[h]
\centering
\caption{LLM model names and their corresponding version identifiers.}
\label{tab:model-versions}
\small
\setlength{\tabcolsep}{6pt}
\begin{tabular}{ll}
\toprule
Model Name & Version Identifier \\
\midrule
\multicolumn{2}{l}{\textit{Main Experiments:}} \\
Gemini-3-Pro & gemini-3-pro-preview-11-2025-thinking \\
\midrule
\multicolumn{2}{l}{\textit{Model Ablation Study:}} \\
GPT-5.1 & gpt-5.1-all \\
Claude-4.5 & claude-sonnet-4-5-20250929 \\
Gemini-3-Pro & gemini-3-pro-preview-11-2025-thinking \\
Gemini-2.5-Pro & gemini-2.5-pro \\
DeepSeek-v3.2 & deepseek-v3.2 \\
Qwen-3-235B & qwen3-235b-a22b-instruct-2507 \\
GLM-4.7 & glm-4.7 \\
Grok-4.1 & grok-4.1 \\
\bottomrule
\end{tabular}
\end{table}

\textbf{Implementation details.}
All models use identical sampling parameters: temperature 0.7, top-p 0.95, and max tokens 4096.
The main experiments (ReAct, Reflexion, DPT-WToM, and CoWork-X baselines) use Gemini-3-Pro-Preview-Thinking exclusively to ensure fair comparison.
The model ablation study (Section 5.3) evaluates CoWork-X across different model families to demonstrate generalizability.

\section{Details of Skill Library}
\label{sec:skill-lib-detail}
\textbf{Original Skill Library.} \cref{fig:origin-skill} illustrates the initial, minimal yet executable skill library. This baseline comprises solely fundamental state representations, function definition for methods and operators, and their corresponding Pyhop declarations. Within the method functions, we provide exemplar subtask lists to prevent the Co-Optimizer from generating logic incompatible with the game environment. For the remaining functions, we supply default state returns accompanied by instructional comments to guide the Co-Optimizer in function completion. Under this original rule set, Gemini-3-pro achieved a score of -60. We designate this infancy state as the initialization point for the CoWork-X optimization process.\\
\textbf{Best-Optimized Skill Library.} \cref{fig:optimized-code} illustrates the optimal skill library derived from the 18th optimization iteration of Gemini-3-pro. Guided by this library, the skill agent attained a peak score of 180. Notably, subsequent iterations achieving this score utilized structurally similar libraries, indicating that the optimization process effectively converged at this point. This library provides comprehensive and precise definitions for all methods and operators, ensuring that the agent executes the game without triggering any runtime errors.

\section{Failure Mode}
\label{sec:fail}

\cref{fig:failure-mode} illustrates an instance of optimization failure exhibited by Qwen-3-225B. Constrained by inherent model capabilities, Qwen-3-225B accumulated numerous erroneous variable definitions—such as beef\_fresh\_count across multiple execution epochs. Within the game environment, the model failed to accurately pinpoint the error locations based on the provided error logs, thereby impeding the Co-Optimizer's effective self-optimization.
\section{License}

The source code will be available under the MIT License. 
The full text of the license is provided at~\url{https://opensource.org/licenses/MIT}.

\begin{figure*}
    \begin{prompt}{Prompt of Co-Optimizer}
You are an expert in HTN (Hierarchical Task Network) planning and Python programming.\\
We are optimizing a rule file for an autonomous agent in an Overcooked-like environment using the Pyhop planner.\\
\{history\_context\}\\
\{best\_rules\_context\}\\

Current Rule File \{current\_rule\_file\}:\\
\{current\_rules\}\\

origin Rule File \{origin\_rule\_file\}:\\
\{origin\_rules\}\\
Notice: Every variant name and function name in origin rule file is correct, although origin rule is broken. Do not change any basic elements in origin rule file, only to complete, optimize the given function and write new process logic of other recipes.\\

Experiment Result (Iteration \{iteration\}):\\
Score: \{score\}

Log Analysis:
\{analysis log\}\\

Task:\\
Analyze the logs and the current rules to identify if the rule is perfect.\\
Common issues include:\\
1.  **Precondition Failures**: Operators returning `False` because state checks fail.\\
2.  **State Update Failures**: Operators not updating the state correctly (e.g., not incrementing counts).\\
3.  **Method Logic**: Methods not decomposing into the correct sequence of operators.\\
4.  **Missing Logic**: Handling for specific ingredients or burgers is missing.\\
5.  **Recipe Errors**: Incorrect ingredient counts (e.g., requiring 2 items when 1 is standard).\\

Your Goal:
Modify the rule file to FIX the errors (if exist) and MAXIMIZE the score.\\
- If you see some recipe logic missing, add it.\\
- If you see "Assemble" failing, check the ingredient requirements and counts.\\
- If you see a method missing a step, add it.\\
- Learn from the optimization history - don't repeat mistakes from previous iterations.\\

CRITICAL INSTRUCTIONS:\\
- Output ONLY valid Python code.\\
- Do NOT include any explanations, markdown formatting, or comments outside the code.\\
- Do NOT start with text like `Looking at the code...' or `Here is the fix...'\\
- The first line MUST be valid Python (import statement, comment, or function definition)\\
- Your entire response will be saved directly as a .py file\\
- If you think the rule is already perfect, output current rules totally the same.\\

Start your response with the first line of Python code immediately.\\
\end{prompt}
    \caption{Prompt of Co-Optimizer.} 
    \label{fig:optimize-prompt}
\end{figure*}

\begin{figure*}
    \begin{prompt}{Original Skill Library}
from pyhop import hop as pyhop\\

\# determine if some ingredient is available. You need to complete the logic corresponding to this function according to the variable name given in the log.\\
def is\_available(state, ingredient):\\
\hspace*{2em} return True\\

\# prepare food and update state. You need to complete the update logic corresponding to this function according to the variable name given in the log.\\
def op\_prepare\_food(state, agent, food\_type):\\
\hspace*{2em} return state\\

\# assemble used ingredient and update state. You need to complete the update logic corresponding to this function according to the variable name given in the log.\\
def op\_assemble(state, agent, burger\_type):\\
\hspace*{2em} return state\\

\# serve finished burger and update state. You need to complete the update logic corresponding to this function according to the variable name given in the log.\\
def op\_serve(state, agent, burger\_type):\\
\hspace*{2em} return state\\

\# return subtasks can only contains functions start with op\_, and end with assemble and serve. do not use recursion\\
def m\_make\_beef\_burger(state, agent):\\
    \hspace*{2em} if getattr(state, `beef\_burger\_count', 0) $>$ 0:\\
        \hspace*{4em} return [(`op\_serve', agent, `BeefBurger')]\\
    \hspace*{2em} if is\_available(state, `Beef'):\\
        \hspace*{4em} return [(`op\_assemble', agent, `BeefBurger'), (`op\_serve', agent, `BeefBurger')]\\
    \hspace*{2em} return False\\

\# declare defined methods and operators\\
def declare\_rules(pyhop\_module):\\
    \hspace*{2em} pyhop\_module.declare\_operators(op\_prepare\_food, op\_assemble, op\_serve)\\
    \hspace*{2em} pyhop\_module.declare\_methods(`make\_BeefBurger', m\_make\_beef\_burger)\\
\end{prompt}
    \caption{Original Skill Library.}
    \label{fig:origin-skill}
\end{figure*}

\begin{figure*}
    \begin{prompt}{Optimized Best Skill Library}
    \scriptsize
from pyhop import hop as pyhop\\
\# Determine if some ingredient is available. Buffer `Bread' ($>$ 1) to mitigate race conditions.\\
def is\_available(state, ingredient):\\
\hspace*{2em}if ingredient == `Beef': return getattr(state, `beef\_cooked\_count', 0) $>$ 0\\
\hspace*{2em}elif ingredient == `Lettuce': return getattr(state, `lettuce\_chopped\_count', 0) $>$ 0\\
\hspace*{2em}elif ingredient == `Bread': return getattr(state, `bread\_count', 0) $>$ 1 \# Require buffer\\
\hspace*{2em}return False\\
\# Prepare food and update state counters.\\
def op\_prepare\_food(state, agent, food\_type):\\
\hspace*{2em}if food\_type == `Beef': state.beef\_cooked\_count = getattr(state, `beef\_cooked\_count', 0) + 1\\
\hspace*{2em}elif food\_type == `Lettuce': state.lettuce\_chopped\_count = getattr(state, `lettuce\_chopped\_count', 0) + 1\\
\hspace*{2em}elif food\_type == `Bread': state.bread\_count = getattr(state, `bread\_count', 0) + 1\\
\hspace*{2em}return state\\
\# Assemble used ingredients and update state.\\
def op\_assemble(state, agent, burger\_type):\\
\hspace*{2em}bread = getattr(state, `bread\_count', 0)\\
\hspace*{2em}if bread $<$ 1: return False\\
\hspace*{2em}if burger\_type == `BeefBurger':\\
\hspace*{4em}beef = getattr(state, `beef\_cooked\_count', 0)\\
\hspace*{4em}if beef $>$= 1: \\
\hspace*{6em}state.bread\_count=bread-1; state.beef\_cooked\_count=beef-1; \\
\hspace*{6em}state.beef\_burger\_count=getattr(state, `beef\_burger\_count',0)+1; \\
\hspace*{6em}return state\\
\hspace*{2em}elif burger\_type == `LettuceBurger':\\
\hspace*{4em}lettuce = getattr(state, `lettuce\_chopped\_count', 0)\\
\hspace*{4em}if lettuce $>$= 1:\\ 
\hspace*{6em}state.bread\_count=bread-1; state.lettuce\_chopped\_count=lettuce-1; \\
\hspace*{6em}state.lettuce\_burger\_count=getattr(state, `lettuce\_burger\_count',0)+1; \\
\hspace*{6em}return state\\
\hspace*{2em}elif burger\_type == `BeefLettuceBurger':\\
\hspace*{4em}beef, lettuce = getattr(state, `beef\_cooked\_count', 0), getattr(state, `lettuce\_chopped\_count', 0)\\
\hspace*{4em}if beef $>$= 1 and lettuce $>$= 1:\\ 
\hspace*{6em}state.bread\_count=bread-1; state.beef\_cooked\_count=beef-1; \\
\hspace*{6em}state.lettuce\_chopped\_count=lettuce-1; \\
\hspace*{6em}state.beef\_lettuce\_burger\_count=getattr(state, `beef\_lettuce\_burger\_count',0)+1; \\
\hspace*{6em}return state\\
\hspace*{2em}return False\\
\# Serve finished burger and update state.\\
def op\_serve(state, agent, burger\_type):\\
\hspace*{2em}if burger\_type == `BeefBurger':\\
\hspace*{4em}if getattr(state, `beef\_burger\_count', 0) $>$ 0: state.beef\_burger\_count -= 1; return state\\
\hspace*{2em}elif burger\_type == `LettuceBurger':\\
\hspace*{4em}if getattr(state, `lettuce\_burger\_count', 0) $>$ 0: state.lettuce\_burger\_count -= 1; return state\\
\hspace*{2em}elif burger\_type == `BeefLettuceBurger':\\
\hspace*{4em}if getattr(state, `beef\_lettuce\_burger\_count', 0) $>$ 0: state.beef\_lettuce\_burger\_count -= 1; return state\\
\hspace*{2em}return False\\
\# Method: Make BeefBurger\\
def m\_make\_beef\_burger(state, agent):\\
\hspace*{2em}if getattr(state, `beef\_burger\_count', 0) $>$ 0: return [(`op\_serve', agent, `BeefBurger')]\\
\hspace*{2em}tasks = []\\
\hspace*{2em}if not is\_available(state, `Beef'): tasks.append((`op\_prepare\_food', agent, `Beef'))\\
\hspace*{2em}if not is\_available(state, `Bread'): tasks.append((`op\_prepare\_food', agent, `Bread'))\\
\hspace*{2em}return tasks + [(`op\_assemble', agent, `BeefBurger'), (`op\_serve', agent, `BeefBurger')]\\
\# Method: Make LettuceBurger\\
def m\_make\_lettuce\_burger(state, agent):\\
\hspace*{2em}if getattr(state, `lettuce\_burger\_count', 0) $>$ 0: return [(`op\_serve', agent, `LettuceBurger')]\\
\hspace*{2em}tasks = []\\
\hspace*{2em}if not is\_available(state, `Lettuce'): tasks.append((`op\_prepare\_food', agent, `Lettuce'))\\
\hspace*{2em}if not is\_available(state, `Bread'): tasks.append((`op\_prepare\_food', agent, `Bread'))\\
\hspace*{2em}return tasks + [(`op\_assemble', agent, `LettuceBurger'), (`op\_serve', agent, `LettuceBurger')]\\
\# Method: Make BeefLettuceBurger (Priority: Beef -$>$ Lettuce -$>$ Bread)\\
def m\_make\_beef\_lettuce\_burger(state, agent):\\
\hspace*{2em}if getattr(state, `beef\_lettuce\_burger\_count', 0) $>$ 0: return [(`op\_serve', agent, `BeefLettuceBurger')]\\
\hspace*{2em}tasks = []\\
\hspace*{2em}if not is\_available(state, `Beef'): tasks.append((`op\_prepare\_food', agent, `Beef'))\\
\hspace*{2em}if not is\_available(state, `Lettuce'): tasks.append((`op\_prepare\_food', agent, `Lettuce'))\\
\hspace*{2em}if not is\_available(state, `Bread'): tasks.append((`op\_prepare\_food', agent, `Bread'))\\
\hspace*{2em}return tasks + [(`op\_assemble', agent, `BeefLettuceBurger'), (`op\_serve', agent, `BeefLettuceBurger')]\\
\# Declare defined methods and operators\\
def declare\_rules(pyhop\_module):\\
\hspace*{2em}pyhop\_module.declare\_operators(op\_prepare\_food, op\_assemble, op\_serve)\\
\hspace*{2em}pyhop\_module.declare\_methods(`make\_BeefBurger', m\_make\_beef\_burger)\\
\hspace*{2em}pyhop\_module.declare\_methods(`make\_LettuceBurger', m\_make\_lettuce\_burger)\\
\hspace*{2em}pyhop\_module.declare\_methods(`make\_BeefLettuceBurger', m\_make\_beef\_lettuce\_burger)\\
\end{prompt}
    \caption{ Best-optimized skill library by Gemini-3-pro.}
    \label{fig:optimized-code}
\end{figure*}

\begin{figure*}
    \begin{prompt}{Failure Mode}
    \scriptsize
from pyhop import hop as pyhop\\
\# Determine if some ingredient is available. Logic based on variable names from logs.\\
def is\_available(state, ingredient):\\
\hspace*{2em}if ingredient == `Beef': return (getattr(state, `beef\_fresh\_count', 0) $>$ 0 or getattr(state, `beef\_in\_progress\_count', 0) $>$ 0 or getattr(state, `beef\_cooked\_count', 0) $>$ 0)\\
\hspace*{2em}elif ingredient == `Lettuce': return (getattr(state, `lettuce\_unchopped\_count', 0) $>$ 0 or getattr(state, `lettuce\_chopped\_count', 0) $>$ 0)\\
\hspace*{2em}elif ingredient == `Bread': return getattr(state, `bread\_count', 0) $>$ 0\\
\hspace*{2em}return False\\
\# Prepare food and update state.\\
def op\_prepare\_food(state, agent, food\_type):\\
\hspace*{2em}if food\_type == `Beef':\\
\hspace*{4em}if getattr(state, `beef\_fresh\_count', 0) $>$ 0: state.beef\_fresh\_count -= 1; state.beef\_in\_progress\_count = getattr(state, `beef\_in\_progress\_count', 0) + 1; return state\\
\hspace*{4em}return False\\
\hspace*{2em}elif food\_type == `Lettuce':\\
\hspace*{4em}if getattr(state, `lettuce\_unchopped\_count', 0) $>$ 0: state.lettuce\_unchopped\_count -= 1; state.lettuce\_chopped\_count = getattr(state, `lettuce\_chopped\_count', 0) + 1; return state\\
\hspace*{4em}return False\\
\hspace*{2em}elif food\_type == `Bread': state.bread\_count = getattr(state, `bread\_count', 0) + 1; return state\\
\hspace*{2em}return False\\
\# Cook beef (in progress -$>$ cooked) \& Chop lettuce (unchopped -$>$ chopped)\\
def op\_cook\_beef(state, agent):\\
\hspace*{2em}if getattr(state, `beef\_in\_progress\_count', 0) $>$ 0: state.beef\_in\_progress\_count -= 1; state.beef\_cooked\_count = getattr(state, `beef\_cooked\_count', 0) + 1; return state\\
\hspace*{2em}return False\\
def op\_chop\_lettuce(state, agent):\\
\hspace*{2em}if getattr(state, `lettuce\_unchopped\_count', 0) $>$ 0: state.lettuce\_unchopped\_count -= 1; state.lettuce\_chopped\_count = getattr(state, `lettuce\_chopped\_count', 0) + 1; return state\\
\hspace*{2em}return False\\
\# Assemble used ingredient and update state.\\
def op\_assemble(state, agent, burger\_type):\\
\hspace*{2em}if burger\_type == `BeefBurger':\\
\hspace*{4em}if getattr(state, `beef\_cooked\_count', 0) $<$ 1 or getattr(state, `bread\_count', 0) $<$ 1: return False\\
\hspace*{4em}state.beef\_burger\_count = getattr(state, `beef\_burger\_count', 0) + 1; state.beef\_cooked\_count -= 1; state.bread\_count -= 1; return state\\
\hspace*{2em}elif burger\_type == `LettuceBurger':\\
\hspace*{4em}if getattr(state, `lettuce\_chopped\_count', 0) $<$ 1 or getattr(state, `bread\_count', 0) $<$ 1: return False\\
\hspace*{4em}state.lettuce\_burger\_count = getattr(state, `lettuce\_burger\_count', 0) + 1; state.lettuce\_chopped\_count -= 1; state.bread\_count -= 1; return state\\
\hspace*{2em}elif burger\_type == `BeefLettuceBurger':\\
\hspace*{4em}if getattr(state, `beef\_cooked\_count', 0) $<$ 1 or getattr(state, `lettuce\_chopped\_count', 0) $<$ 1 or getattr(state, `bread\_count', 0) $<$ 1: return False\\
\hspace*{4em}state.beef\_lettuce\_burger\_count=getattr(state, `beef\_lettuce\_burger\_count',0)+1; state.beef\_cooked\_count-=1; state.lettuce\_chopped\_count-=1; state.bread\_count-=1; return state\\
\hspace*{2em}return False\\
def op\_serve(state, agent, burger\_type):\\
\hspace*{2em}if burger\_type == `BeefLettuceBurger' and getattr(state, `beef\_lettuce\_burger\_count', 0) $>$ 0:\\
\hspace*{4em}state.beef\_lettuce\_burger\_count -= 1; state.burger\_count = getattr(state, `burger\_count', 0) + 1; state.total\_score += 15; return state\\
\hspace*{2em}elif burger\_type == `BeefBurger' and getattr(state, `beef\_burger\_count', 0) $>$ 0:\\
\hspace*{4em}state.beef\_burger\_count -= 1; state.burger\_count = getattr(state, `burger\_count', 0) + 1; state.total\_score += 10; return state\\
\hspace*{2em}elif burger\_type == `LettuceBurger' and getattr(state, `lettuce\_burger\_count', 0) $>$ 0:\\
\hspace*{4em}state.lettuce\_burger\_count -= 1; state.burger\_count = getattr(state, `burger\_count', 0) + 1; state.total\_score += 10; return state\\
\hspace*{2em}return False\\
def m\_make\_beef\_burger(state, agent):\\
\hspace*{2em}if getattr(state, `beef\_burger\_count', 0) $>$ 0: return [(`op\_serve', agent, `BeefBurger')]\\
\hspace*{2em}if is\_available(state, `Beef') and is\_available(state, `Bread'):\\
\hspace*{4em}return [(`op\_prepare\_food', agent, `Beef'), (`op\_cook\_beef', agent), (`op\_prepare\_food', agent, `Bread'), (`op\_assemble', agent, `BeefBurger'), (`op\_serve', agent, `BeefBurger')]\\
\hspace*{2em}return False\\
def m\_make\_lettuce\_burger(state, agent):\\
\hspace*{2em}if getattr(state, `lettuce\_burger\_count', 0) $>$ 0: return [(`op\_serve', agent, `LettuceBurger')]\\
\hspace*{2em}if is\_available(state, `Lettuce') and is\_available(state, `Bread'):\\
\hspace*{4em}return [(`op\_prepare\_food', agent, `Lettuce'), (`op\_chop\_lettuce', agent), (`op\_prepare\_food', agent, `Bread'), (`op\_assemble', agent, `LettuceBurger'), (`op\_serve', agent, `LettuceBurger')]\\
\hspace*{2em}return False\\
def m\_make\_beef\_lettuce\_burger(state, agent):\\
\hspace*{2em}if getattr(state, `beef\_lettuce\_burger\_count', 0) $>$ 0: return [(`op\_serve', agent, `BeefLettuceBurger')]\\
\hspace*{2em}if is\_available(state, `Beef') and is\_available(state, `Lettuce') and is\_available(state, `Bread'):\\
\hspace*{4em}return [(`op\_prepare\_food', agent, `Beef'), (`op\_cook\_beef', agent), (`op\_prepare\_food', agent, `Lettuce'), (`op\_chop\_lettuce', agent), (`op\_prepare\_food', agent, `Bread'), (`op\_assemble', agent, `BeefLettuceBurger'), (`op\_serve', agent, `BeefLettuceBurger')]\\
\hspace*{2em}return False\\
def declare\_rules(pyhop\_module):\\
\hspace*{2em}pyhop\_module.declare\_operators(op\_prepare\_food, op\_cook\_beef, op\_chop\_lettuce, op\_assemble, op\_serve)\\
\hspace*{2em}pyhop\_module.declare\_methods(`make\_BeefBurger', m\_make\_beef\_burger)\\
\hspace*{2em}pyhop\_module.declare\_methods(`make\_LettuceBurger', m\_make\_lettuce\_burger)\\
\hspace*{2em}pyhop\_module.declare\_methods(`make\_BeefLettuceBurger', m\_make\_beef\_lettuce\_burger)\\
\end{prompt}
    \caption{A failed skill library in runs of Qwen-3-225B.}
    \label{fig:failure-mode}
\end{figure*}

\end{document}